\documentclass[10pt,twocolumn,letterpaper]{article}

\usepackage[pagenumbers]{cvpr} % To force page numbers, e.g. for an arXiv version

\usepackage{graphicx}
\usepackage{amsmath}
\usepackage{amssymb}
\usepackage{booktabs}
\usepackage{comment}
\usepackage{multirow}
\usepackage{float}% \usepackage{pgfplots}
\usepackage{enumitem}% http://ctan.org/pkg/enumitem
\usepackage{tablefootnote}

\usepackage[pagebackref,breaklinks,colorlinks]{hyperref}

\usepackage[capitalize]{cleveref}
\crefname{section}{Sec.}{Secs.}
\Crefname{section}{Section}{Sections}
\Crefname{table}{Table}{Tables}
\crefname{table}{Tab.}{Tabs.}

\begin{document}

\title{Building Inspection Toolkit: Unified Evaluation and Strong Baselines for Damage Recognition}

\author{Johannes Flotzinger, Philipp J. Rösch, Norbert Oswald, Thomas Braml\\
University of the Bundeswehr Munich\\
{\tt\small \{johannes.flotzinger,philipp.roesch,norbert.oswald,thomas.braml\}@unibw.de}
%\and
}
\maketitle

%%%%%%%%% ABSTRACT
\begin{abstract}
In recent years, several companies and researchers have started to tackle the problem of damage recognition within the scope of automated inspection of built structures. 
While companies are  neither willing to publish associated data nor models, researchers are facing the problem of data shortage on one hand and inconsistent dataset splitting with the absence of consistent metrics on the other hand. This leads to incomparable results. Therefore, we introduce the building inspection toolkit -- bikit -- which acts as a simple to use data hub containing relevant open-source datasets in the field of damage recognition. The datasets are enriched with evaluation splits and predefined metrics, suiting the specific task and their data distribution. For the sake of compatibility and to motivate researchers in this domain, we also provide a leaderboard and the possibility to share model weights with the community. 
As starting point we provide strong baselines for multi-target classification tasks utilizing extensive hyperparameter search using three transfer learning approaches for state-of-the-art algorithms.
The toolkit\footnote{\href{https://github.com/phiyodr/building-inspection-toolkit}{https://github.com/phiyodr/building-inspection-toolkit}} and the leaderboard\footnote{\href{https://dacl.ai}{https://dacl.ai}} are available online.

\end{abstract}

%%%%%%%%% BODY TEXT
%-------------------------------------------------------------------------
\section{Introduction}
\label{sec:intro}

% -------------------------------------------------------------------------
\begin{figure}[t]
    \centering
    \includegraphics[height=0.125\textheight]{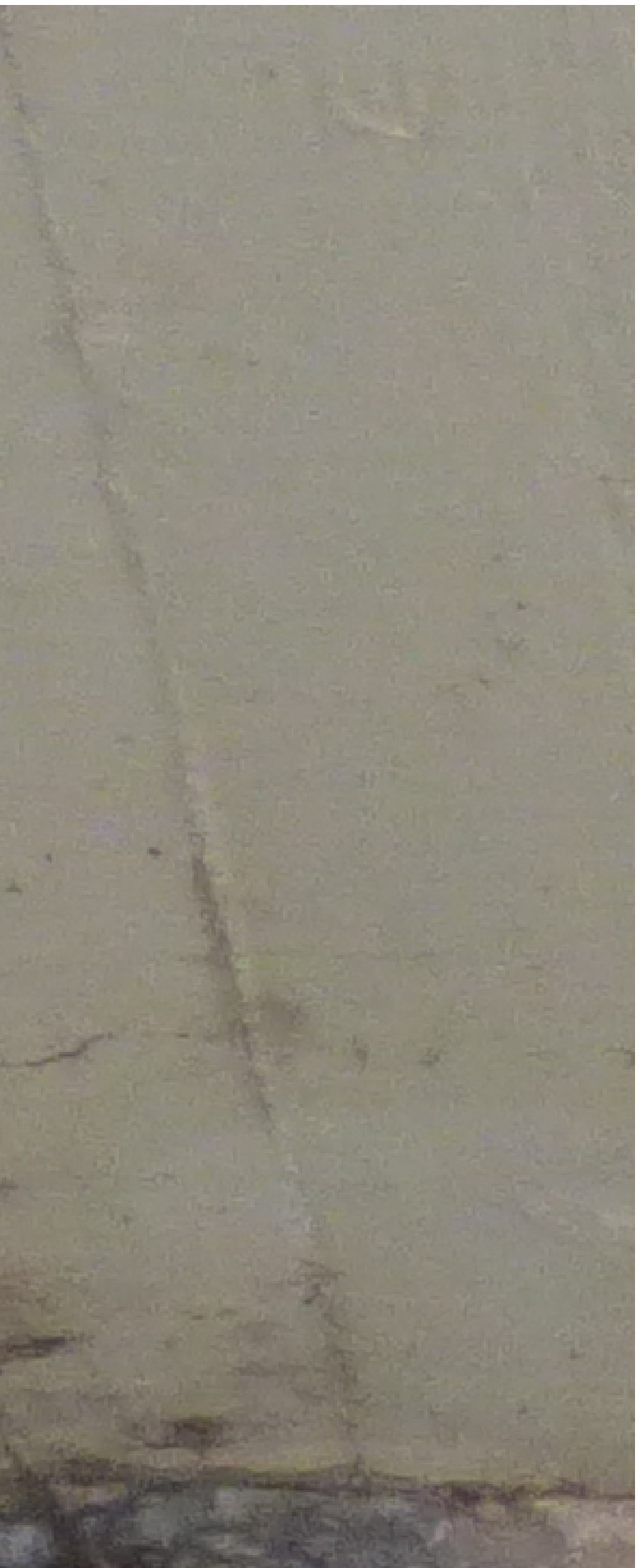}\hfill
    \includegraphics[height=0.125\textheight]{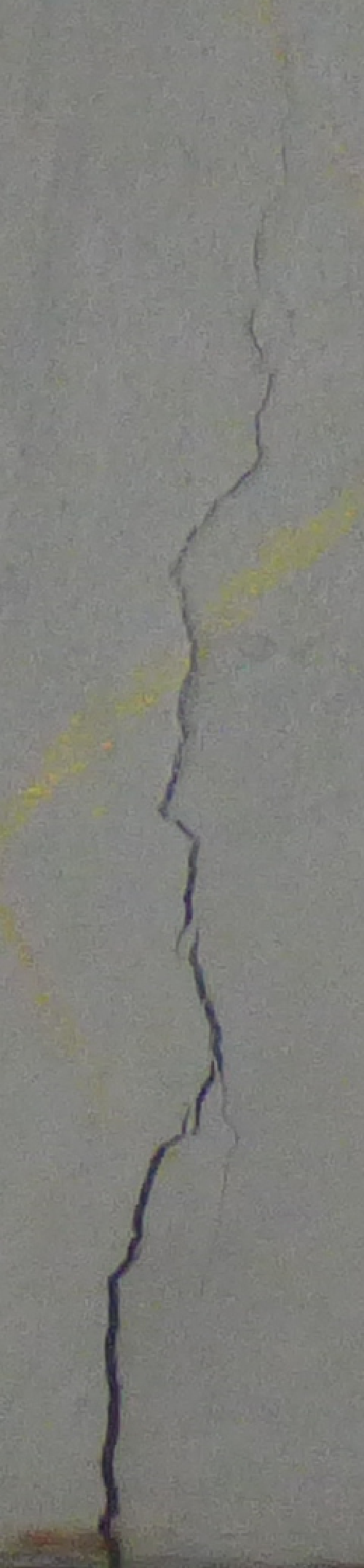}\hfill
    \includegraphics[height=0.125\textheight]{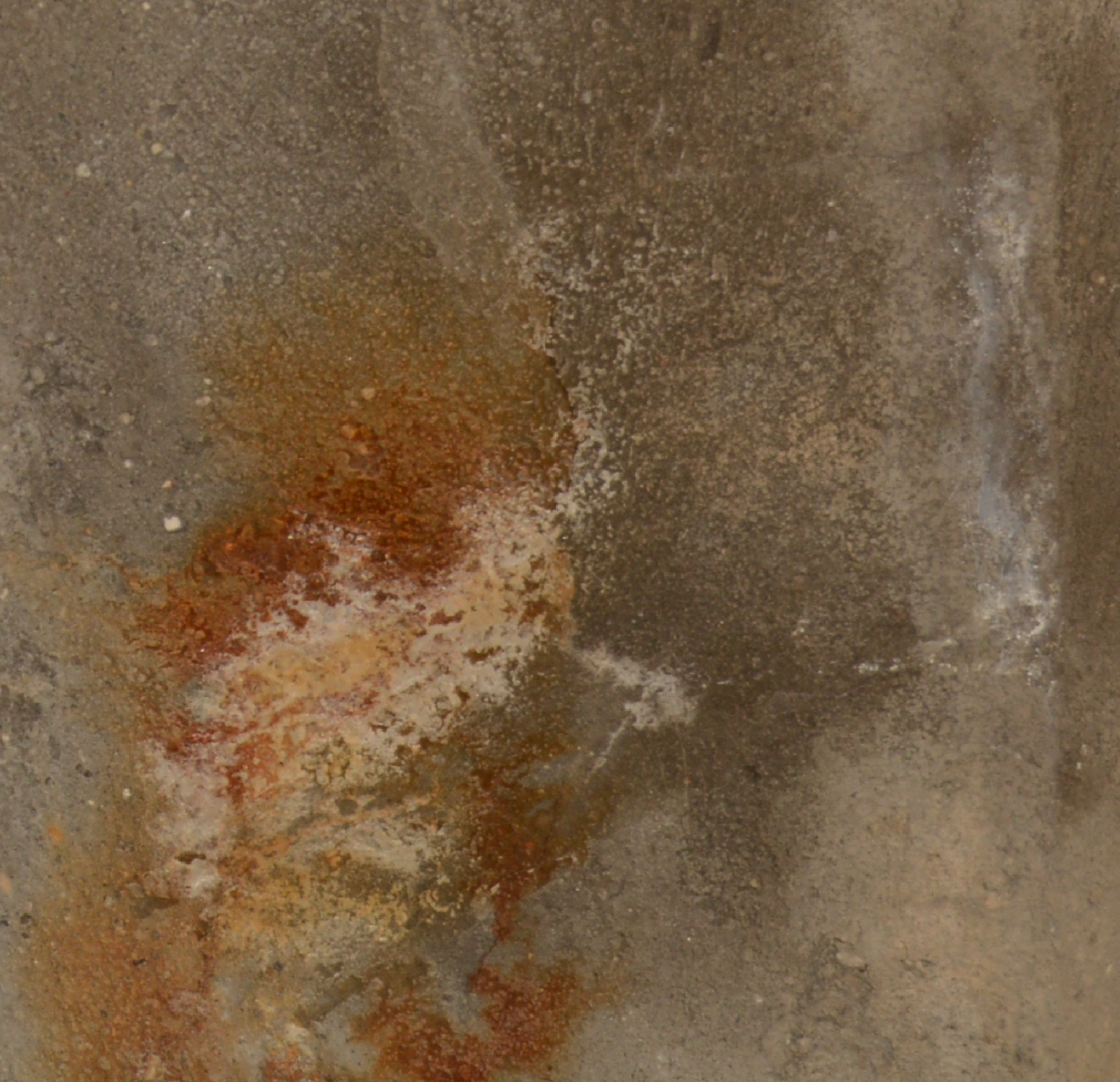}\hfill
    \includegraphics[height=0.125\textheight]{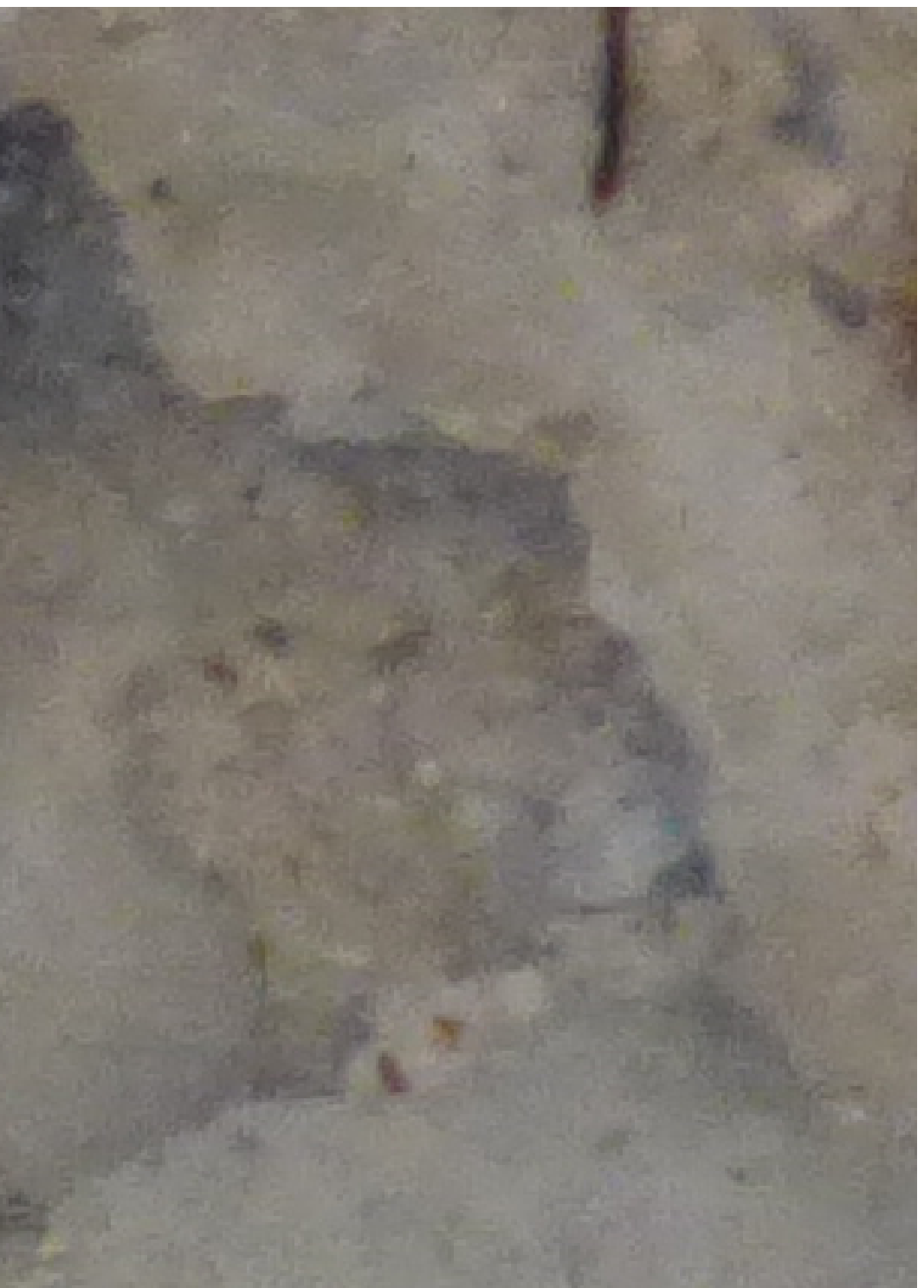}\hfill
    \includegraphics[height=0.125\textheight]{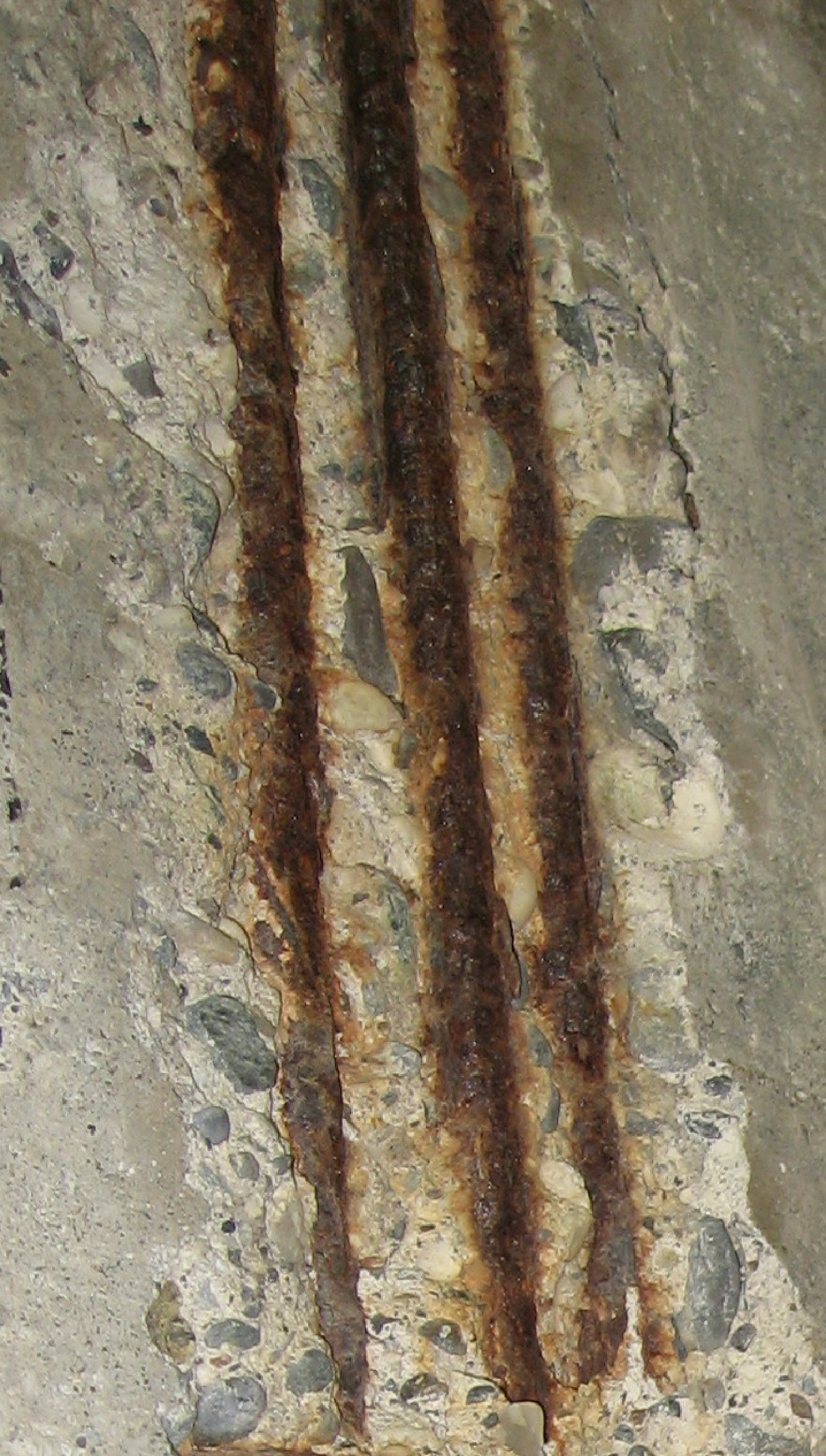}\hfill
    \\[\smallskipamount]
    \includegraphics[height=0.0897\textheight]{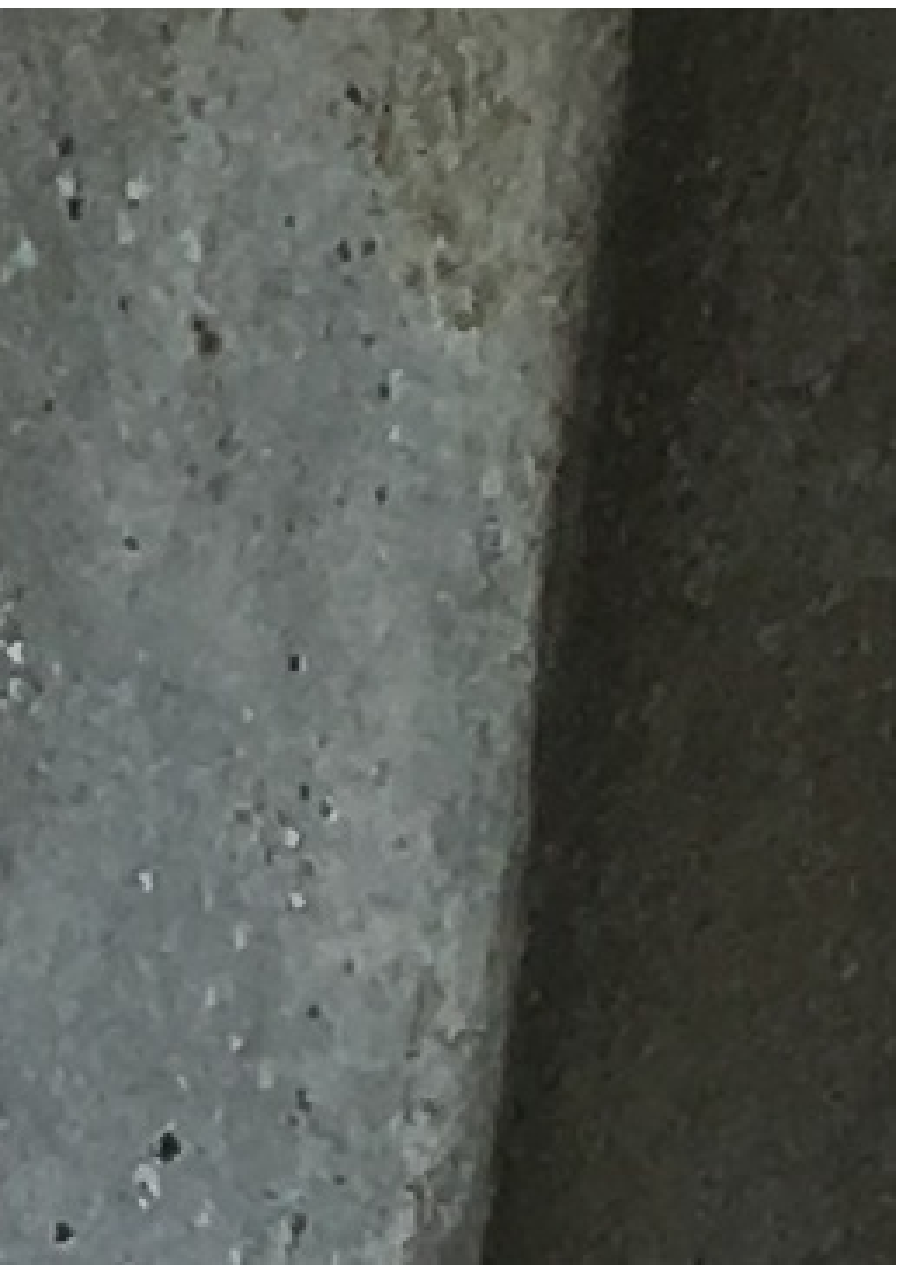}\hfill
    \includegraphics[height=0.0897\textheight]{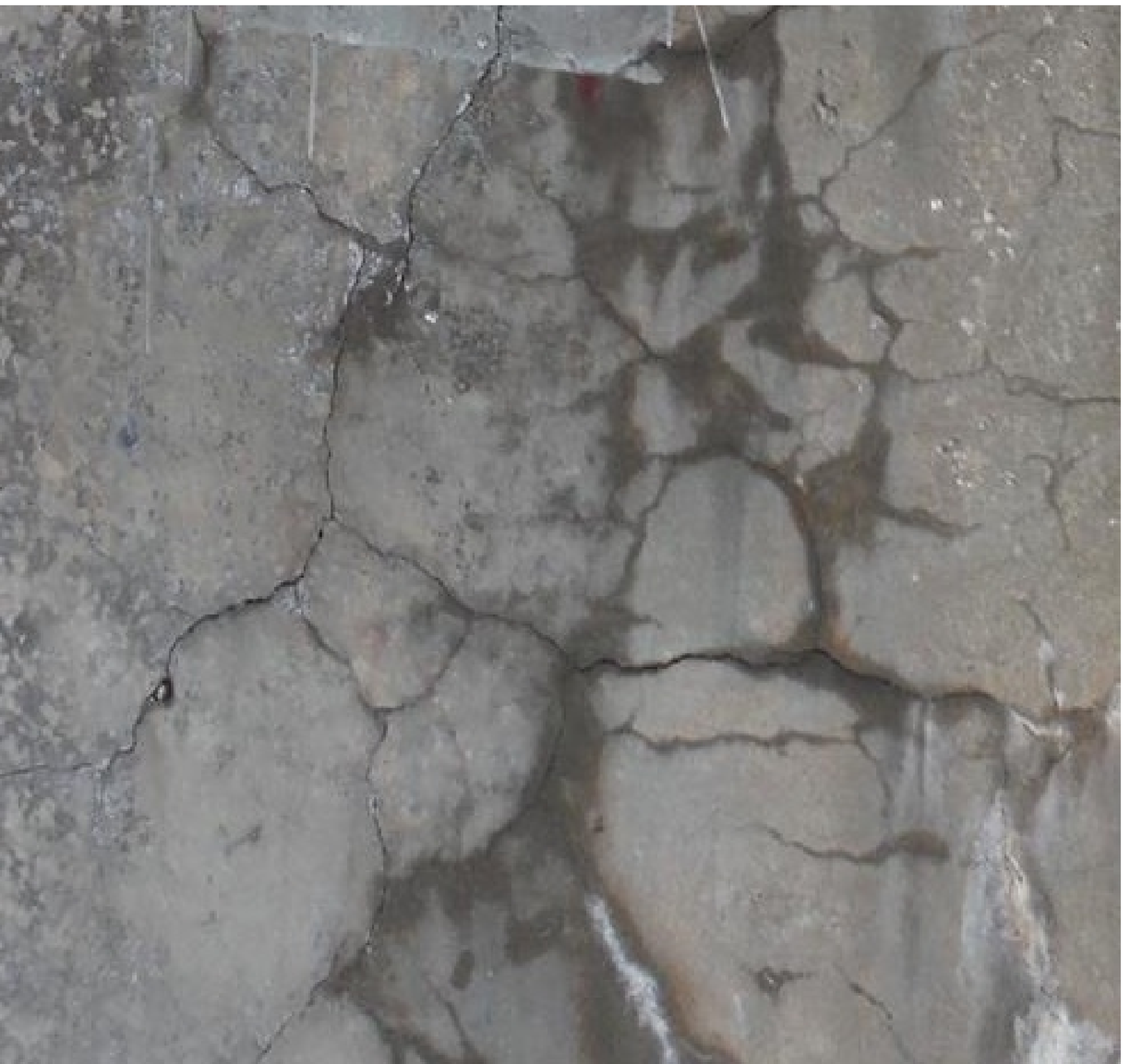}\hfill
    \includegraphics[height=0.0897\textheight]{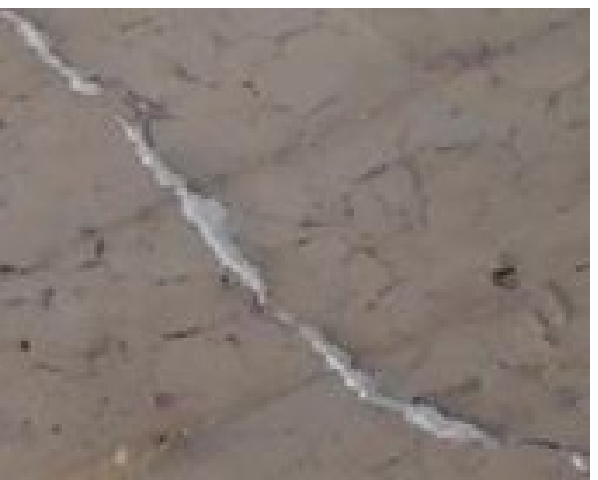}\hfill
    \includegraphics[height=0.0897\textheight]{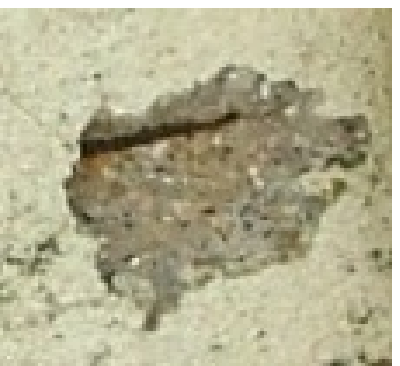}\hfill
    \label{fig:CommonDamages}
\caption{Example damages from CODEBRIM (top line) and MCDS (bottom line). 
The labels for CODEBRIM are No Damage, Crack, Efflorescence/Rust, Spalling, Spalling/Exposed Bars/Rust. 
The labels for MCDS     are No Damage, Crack, Efflorescence, Spalling/Exposed Bars/Rust.}
\end{figure}
% -------------------------------------------------------------------------

Bridges represent a very important part of the infrastructure worldwide. 
Their closure, reconstruction or abrupt failure due to poor maintenance leads to economical damage or -- even worse -- the loss of human lifes, what happened in Genoa, Italy in 2018 \cite{pianigiani_poor_2020}. 
Therefore, the availability of functioning and safe bridges is crucial. 
Due to the fact that many bridges are currently reaching ages at which damages occur more often \cite{ARTBA2021, Haslbeck2019} 
the assessment intervals must be set more frequently, which increases the pressure  on individual inspectors.

Inspectors usually perform a visual assessment from a short distance to the structure taking pictures and notes of every damage, which is very time consuming. Moreover the classification of the damages heavily depends on the inspector's personal impression. Thereby, the detected damage classifications and recommended consequences in the final reports of different inspectors validating the same building can differ \cite{Phares2004}.
Due to the above mentioned points, an automated damage classification would bring great benefits, like
    (\textit{i}) saving time and lowering the costs for the assessment process,
    (\textit{ii}) less subjective damage classification,
    (\textit{iii}) no need of additional machinery (e.g. mobile under-bridge unit) if UAVs are used to gather the image data, and
    (\textit{iv}) the ability to execute routine inspections by less experienced employees.

We want to accelerate the development in the context of damage recognition and therefore introduce \textit{bikit}. 
It is designed to be an easy to use platform and consists of three components: (\textit{i}) datasets with fixed splits, (\textit{ii}) a public leaderboard with predefined metrics, and (\textit{iii}) the availability to upload trained models. 
First, \textit{bikit} will make all known datasets available with a simple API.
Although being a niche area, there is a reasonable number of datasets for reinforced concrete damages (RCD).  
Most of them are binary classification datasets that examine the presence of cracks \cite{Xu2019, Li2019} or damages in general \cite{Huthwohl2018a}. 
Besides that, there are also more detailed and realistic multi-target dataset \cite{Huthwohl2019, Mundt2019} available. 
In the current \textit{bikit} version we provide all open-source  datasets.
Apart from accessibility, comparability is another crucial feature. 
Hence, fixed test sets and useful metrics are required.  
We supply predefined training, validation, and testing splits if missing.
On the basis of data distribution, we create individual splits. 
Moreover, useful metrics are defined for each dataset for performance evaluation. 
All datasets are provided with a strong baseline using state-of-the-art algorithms. For this purpose, we run an expensive hyperparameter search using three different transfer learning strategies on famous (pre-trained) models and make these available online.

%-------------------------------------------------------------------------
\section{Similar Approaches}
\label{sec:RelatedWork}

\textbf{Data and Model Hubs.}
Machine learning research is based on the availability of data. In recent years, the number of datasets that are open to the public has continued to grow and the usage has been simplified. Established standard data for various domains is directly accessible via TensorFlow or PyTorch modules. 
However, specialized hubs have developed as well. For example, \textit{datasets} \cite{lhoest2021datasets} is a platform for a large number of NLP datasets from a wide variety of fields. Besides renowned datasets -- often cited in research -- niche datasets and low-resource languages can also be accessed.
\textit{OpenML} \cite{OpenML2013} provides metrics and leaderboards in addition to data and helps not only with readily available data but also with results on specific hyperparameter settings. In terms of supply of metrics, the same is true for FAIR's \textit{paperswithcode.com} \cite{paperswithcode}.
Moreover, computer vision platforms like \textit{timm} \cite{rw2019timm} do not only provide high-quality code for established models but also model weights to simplify usage and accessibility. 
The above mentioned developments accelerate research and make results more comparable.

\textbf{Transfer Learning.}
The most comprehensive work in the area of RCD recognition was executed by Bukhsh \textit{et al.} \cite{Bukhsh2021}. They investigated several concepts of transfer learning for six classification datasets. Transfer learning strategies consist of varying combinations of upstream and downstream datasets using cross-domain and in-domain data. Hence, training from scratch has been evaluated as well. 
Results show that in-domain and cross-domain perform almost identically well for small datasets, but training on both data types together improves the performance. 
In their analysis, common CNNs that have been developed before the year 2015, were taken into account. 
We decided to use more recent architectures for our analysis. 
Moreover, they only trained the networks' head. 
Besides that, we also investigate the gain of updating the whole model.

\section{Building Inspection Toolkit}

The toolkit consists of three parts, which are elaborated in more detail in the following sections. 
The first building block are the datasets conceived by other researchers and now harmonized into one module \cite{rosch2022bikit}. 
We also add training, validation and testing data splits, if they do not exist. The second component is the introduction of performance metrics, which fit task type and data distribution. 
Results are made available via an online leaderboard. 
As a third part we launch a model hub to make models accessible.

\subsection{Datasets}
\label{sec:Datasets}

\begin{table}
  \centering
  \begin{tabular}{l r r r r r r}
    \toprule
    Data  & Classes & Types   & Size & Splits      \\ 
	\midrule 
  	CDS \cite{Huthwohl2018a} & 2 & STC & 1,028 & no  \\ 
	BCD \cite{Xu2019}& 2 &  STC & 5,390 & \checkmark  \\ 
  	SDNET2018 \cite{Dorafshan2018}& 2 & STC & 13,620 & no   \\ 
   	ICCD\tablefootnote{Data is only available upon request.} \cite{Li2019} & 2 & STC & 60,010 	& \checkmark  \\ 
   	MCDS \cite{Huthwohl2019} & 8 & MTC &  3,607 & no  \\ 
   	CODEBRIM \cite{Mundt2019} & 6 & MTC &  7,261 & \checkmark   \\
    \bottomrule
  \end{tabular}
  \caption{Open-source datasets for damage recognition on reinforced concrete bridges. Types are single-target (STC) and multi-target classification (MTC). 
  Splits refer to the presence of fix training, validation and testing splits in the original publication.}
  \label{tab:datasets}
\end{table}

Several RCD datasets are publicly available for single-target and multi-target classification (see \cref{tab:datasets}). 
To our knowledge no open-source object detection and semantic segmentation datasets exist currently. 
In this publication we deal with the two multi-target classification datasets MCDS \cite{Huthwohl2019} and CODEBRIM \cite{Mundt2019}. 
They are more challenging than single-target classification tasks and also more realistic, since several damages can be present on one image at the same time. Moreover, images are taken in a less standardized setting with different camera angles and image resolutions.

In total, there are seven specific damage types, that are currently available in RCD datasets, i.e. \textit{cracks,
efflorescence,
spalling/spallation,
exposed bars/reinforcement,
rust staining/corrosion strain,
scaling}, and 
\textit{other/general defects}. Other types exist, but are aggregated in meta classes. 
CDS \cite{Huthwohl2018a} summarizes \textit{graffiti}, \textit{vegetation}, and \textit{blistering} within the label ``unhealthy''. 
MCDS subordinates \textit{graffiti} and \textit{vegetation} to ``other damages''. 
So far MCDS is the most diverse dataset available, whereas the \textit{German guideline for uniform acquisition, assessment, recording and evaluation of results of structural inspections} \cite{pruf} currently distinguishes approximately 50 damages in total. To conclude, there is still a lot of work to be done, until machine learning-based systems can reliably support civil engineers. 

\textbf{MCDS.} 
This dataset \cite{Huthwohl2019} was created for a three-stage approach where, depending on the results of the first model, a second model may be applied and, depending on that, a third. This dataset was recently transformed into a single-stage approach \cite{Bukhsh2021}, whereby data acquisition procedure was neglected. The second and third stages determine, whether \textit{exposed bars} and \textit{rust staining} is visible. The negative samples have been drawn from stage-one dataset. 
Hence, only a few images without \textit{exposed bars} had the label ``no exposed bars''. 
The same applies for ``no rust''. We cleaned the dataset and transferred it to an eight-class dataset. 
Due to the fact that no data splits were provided we introduced fix splits for training (2057 samples), validation (270), and testing (270).

\textbf{CODEBRIM.}
This dataset is the second richest in terms of damage categories. 
It is missing two classes in comparison to MCDS but comes with a much higher number of unique images. 
The authors also made splits available so that all former research is comparable and no splits need to be created. 
The numbers for each split are 6013 (training), 616  (validation), and 632 (testing). 
We use the balanced version for our experiments.

\begin{table}
  \centering
  \begin{tabular}{lrr}
    \toprule
	Classes & MCDS & CODEBRIM \\  
    \midrule 
	No Damage &  452 & 2,506 \\ 
    Crack & 787 & 2,507  \\
    Efflorescence & 304 & 833 \\  
    Spalling & 422 &  1,898\\
    Exposed Bars &  221 & 1,507 \\
    Rust & 350 & 1,559 \\
    Scaling &  163 & - \\
    Other damages & 264 &  - \\ 
    \midrule
    Number of images & 2,597 & 7,261 \\
    Avg. number of classes/image & 1.14 & 1.13 \\
    \bottomrule
  \end{tabular}
  \caption{Counts of classes in  datasets with additional statistics.}
  \label{tab:dataset-labels}
\end{table}

\subsection{Metrics}
\label{sec:metrics}

The authors of MCDS decided to provide a huge number of metrics. They show 10 different metrics for all three stages and for all classes, which is exhausting.
On the other hand CODEBRIM authors chose to report Exact Match Ratio (EMR) only. Later analysis \cite{Bukhsh2021} chose AUROC,
accuracy,
F1-score,
precision, and  
recall on an aggregated level (not on class level). 
We go a middle course between one overall metric and 10 metrics for each class. 
The main metric for multi-target classification in \textit{bikit} is \textit{EMR}.
Besides that we provide \textit{recall by class}. We make both metrics accessible in an online leaderboard.%\footnote{To be ranked on the leaderboard, please follow the instructions on \href{https://dacl.ai/submit}{https://dacl.ai/submit}.} 

\textbf{EMR.} EMR is a challenging task since all predictions must match. 
It is, for example, relatively simple to determine cracks in comparison to other classes. 
Moreover, crack is the class with the highest occurrence (see \cref{tab:dataset-labels}). 
From this follows that metrics which weight classes equally or, worse, by occurrence are biased towards easily determinable classes, like cracks. 
This makes some metrics appear better than they are. 
While EMR is not a metric created directly for unbalanced data, it is complex enough so that the distribution problem is adequately addressed.
Moreover, a side effect is that current values are still not close to perfect fit, which still leaves room for improvement within this research domain.

\textbf{Recall by class.}
The previous metric was primarily selected from a machine learning perspective.
However, when applied in the real world other metrics are of importance. 
For civil engineers it is mainly relevant to see how many damages are overlooked. In current approaches, machine learning-based software is made to support and not to replace engineers. 
Consequently, it is more vital to detect all damages which exist than missing risky damages. Due to that fact, we decided to provide \textit{recall} measures for all classes, in order to be able to see where current models still fail often.

\subsection{Model Hub}
\label{sec:modelhub}

Besides providing a data API and performance results on a leaderboard, we also create a model hub to accelerate research. Having pre-trained models available does not only help to reproduce results but also to obviate the need to fine-tune models again and again. 
This is especially interesting if researchers want to adapt pre-trained classification models  for later use in object detectors or for semantic segmentation tasks. 
We provide strong baseline models for both datasets elaborated in the next chapter.

%-------------------------------------------------------------------------
\section{Strong Baselines for Damage Recognition}
\label{sec:Baselines}

To fill the model hubs with strong baselines we use three modern CNN architectures and conduct an expensive hyperparameter search with three different transfer learning strategies. 

\textbf{Architectures.} The used CNNs are ResNet50 (RN) \cite{he2015deep}, MobileNetV3-Large (MN) \cite{MobileNet}, and EfficientNetV1-B0 (EN) \cite{EfficientNet}, which all belong to a set of state-of-the-art models. 
We initialize the models with weights from  ImageNet \cite{ImageNet} training. The original heads are removed and replaced by two fully connected layers that have dropout applied before each layer. 

\textbf{Transfer Learning and Hyperparameter Search.}
In total we want to train and compare three different transfer learning strategies, which are described in detail in Section~\ref{sec:Approaches}. For computational reasons, we only will run the extensive hyperparameter search on the simplest strategy and then transfer knowledge to the other approaches. % (i.e \textit{HTA}, \textit{DHB}). 
This is elaborated in Section~\ref{sec:HyperparameterSearch}.

%--------------------------------------------------------------------------------

%-------
\begin{table}
    \centering
    \begin{tabular}{lr} 
    \toprule
    Hyperparam.       & Hyperparameter space                              \\ \midrule
    Hidden layer    & 16, 32, 64, 128, 256, 512, 1024           \\
    Batch size      & 16, 32 (MCDS); 64, 128, 256, 512 (both)      \\
    Learning rate   & 1e-4,  5e-4, 1e-3, 5e-3, 1e-2   \\
    Scheduler       & Constant (CtW), Cosine (CeW) w/ warmup    \\
    Dropout       & 0, 0.1, 0.2, 0.3, 0.4                 \\
    Weight decay    & 1e-7, 1e-6, 1e-5                \\
    \bottomrule
    \end{tabular}
    \caption{Values for the initial parameter search (\textit{HO}) for MCDS and CODEBRIM. The best hyperparameter settings are displayed in \cref{tab:BestHOSetting}.}
    \label{tab:HytuHO}
\end{table}

%-------------------------------------------------------------------------
\subsection{Transfer Learning Strategies}
\label{sec:Approaches}

\begin{figure}[t]
    \centering
    \includegraphics[width=.95\linewidth]{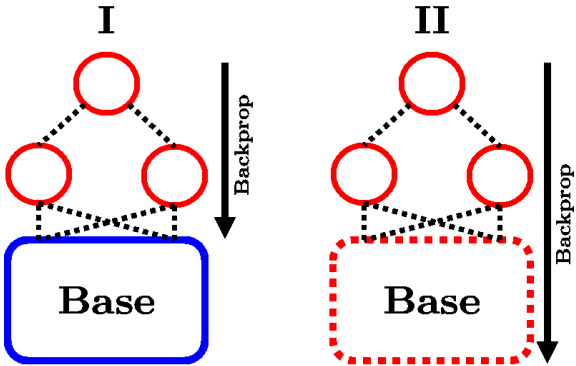}
    \caption{Descriptions of all three transfer learning approaches, where dashed lines indicated trainable weights. \textit{Heads only} (HO) equals the visualization I on the left, \textit{Heads, Then All} (HTA) approach is sequential usage of visualization I and II, and \textit{Different learning rates for Head and Base} (DHB) follows visualization II but with two learning rates.}
    \label{fig:StepsApproaches}
\end{figure}

We use one standard approach and two more advanced strategies. The first only updates the head's weights (\textit{HO}), the second initially trains the head and then all parameters (\textit{HTA}), the last applies different learning rates for model head and base (\textit{DHB}). See \cref{fig:StepsApproaches} for a visualization.

\textbf{HO.}
During the training process, the layers of the feature extractor are frozen, thus only the parameters of the head are updated. The fully-connected layers can be interpreted as a classifier on bottleneck features \cite{Chatfield2014} derived from training on ImageNet objects. This kind of fine-tuning represents a very common way to use learned filters for other datasets. Nevertheless, filters cannot be adjusted for new shapes and structures in RCD datasets.

\textbf{HTA.}
Besides the above mentioned classical method, the whole model can be fine-tuned as well. Kornblith \textit{et al.} \cite{Kornblith2019} showed that this can leverage performance significantly. HTA is structured as a two stage approach. First, only the head is updated until validation loss is saturated. In the second step, all parameters are trainable. Thereby, not only the final classifier but also the filters within the base model can be adjusted towards the data structure of the RCD datasets. 

\textbf{DHB.} 
Instead of consecutively training the head and the entire network afterwards, this approach jointly trains both parts.
This is done by applying different learning rates for base and head which ensures that base and head can adapt at varying pace. By doing so, the learned filters will be adjusted more slightly than the head's parameters. 
This strategy was introduced by Howard and Ruder \cite{Howard2018} as discriminative fine-tuning.

%===========================================
\subsection{Hyperparameter Search}
\label{sec:HyperparameterSearch}

We defined a rich set of parameters, which are investigated in our hyperparameter approach for \textit{HO}. Results are used to reduce the parameter space for two other approaches. All models are trained with comprehensive data augmentation (see Appendix~\ref{app:DataAugmentation}) and using Adam optimizer with weight decay.

\begin{table}
    \centering
    \begin{tabular}{llrrr}
    \toprule
    Data & Hyperparameters &  RN    & EN    & MN  \\
                                      
    \midrule
    CODEBRIM & Hidden layer & 128 & 256 & 1024  \\
            & Batch size & 256  & 256  & 64\\
            & Learning rate & 1e-5 & 1e-7&  1e-6\\
            & Scheduler & CtW & CeW &  CeW\\
            & Dropout          & 0.2  & 0.4  & 0.3      \\
            & Weight decay & 1e-5 & 1e-7 & 1e-6 \\

    \midrule
    MCDS & Hidden layer & 32 & 32 & 64 \\ 
    & Batch size & 64 & 32 & 64 \\
    & Learning rate & 5e-3 &  5e-3 &  5e-3\\
    & Scheduler & CeW & CtW & CtW \\
    & Dropout     & 0.0  & 0.4  & 0.2         \\
    & Weight decay & 1e-5 &  1e-7 &  1e-6  \\
    \bottomrule
    %\multicolumn{11}{l}{*Lr=Learning rate}
    \end{tabular}
    \caption{Best hyperparameter results for \textit{HO} from Bayesian Search for ResNet (RN), EfficientNet (EN), and MobileNet (MN). Possible settings and abbreviations are listed in ~\cref{tab:HytuHO}.}
    \label{tab:BestHOSetting}
\end{table}

%--------------------------------------------------------------------------------

\begin{table}
    \centering
    \begin{tabular}{lr} 
    \toprule
    Hyperparameters       & Learning rate space                          \\ 
    \midrule
    HTA/DHB head   &  1e-4,  5e-4, 1e-3, 5e-3, 1e-2            \\
    DHB  base &        1e-7, 1e-6, 1e-5   \\
    \bottomrule
    \end{tabular}
    \caption{Values for \textit{HTA} and \textit{DHB} approach which are examined for all architectures and datasets. For the remaining hyperparameter see \cref{tab:BestHOSetting}}
    \label{tab:HTA_DHB}
\end{table}

\textbf{HO.}
Training only the head of a pre-trained model is the most straightforward approach for transfer learning. 
Hence, we apply the large hyperparameter space to both datasets and all three CNN models. 
We test different sizes of the hidden layer and various dropout rates. 
Moreover, we experiment with weight decay for Adam and different batch sizes, learning rates, and schedulers (with fix 10 percent warmup steps). 
The settings are listed in \cref{tab:HytuHO}. 
To conduct the hyperparameter optimization \textit{Bayesian Search} \cite{Rasmussen2006, wandb} and \textit{Hyperband-Early-Stopping} \cite{Li2018, wandb} is applied, whereby the maximum number of epochs is set to 100. 
Results of the best settings are provided in ~\cref{tab:BestHOSetting}.
Using the settings from the best tuning result, we rerun the model 5 times with different seeds to verify if results are stable. These results are visualized in ~\cref{fig:ExactMatchRatioCODE} for CODEBRIM and in ~\cref{fig:ExactMatchRatioMCDS} for MCDS.

\textbf{HTA and DHB.}
For \textit{HTA} and \textit{DHB} we adapt the results from \cref{tab:BestHOSetting} and only test different learning rates as indicated in \cref{tab:HTA_DHB}. 
For \textit{HTA} we use the best checkpoint from \textit{HO} for the first stage and then train the model for another 100 epochs but with no frozen weights. Apart from the learning rates, \textit{DHB} uses the best hyperparameter settings from \textit{HO} and is trained for 100 epochs.

%--------------------------------------------------------------------------------
\begin{table}
    \centering
    \begin{tabular}{llrrr}
    \toprule
    Data & Learning rate    &  RN    & EN    & MN  \\
    \midrule
    CODEBRIM & HTA    & 1e-5 & 1e-5 & 1e-5  \\
            & DHB   head & 1e-4 & 1e-3 & 1e-3 \\
            & DHB  base  & 1e-5 & 1e-5 & 1e-5 \\

    \midrule
    MCDS & HTA    & 1e-7 & 1e-5 & 1e-5  \\
            & DHB   head & 1e-4 & 1e-4 & 5e-3 \\
            & DHB  base  & 1e-5 & 1e-5 & 1e-5 \\

    \bottomrule
    \end{tabular}
    \caption{Best learning rate results for \textit{HTA} and \textit{DHB} for ResNet (RN), EfficientNet (EN), and MobileNet (MN). Possible learning rate settings are listed in ~\cref{tab:HTA_DHB}.}
    \label{tab:BestHTA_DHBSetting}
\end{table}

%-------------------------------------------------------------------------
\section{Results of Hyperparameter-Tuning and Transfer Learning Analysis}
\label{sec:ResultsAndDiscussion}

\begin{table*}[h]
    \centering
    \begin{tabular}{lllrrrrrrrrr} 
    \toprule
    \multirow{2}{*}{Data} & \multirow{2}{*}{Approach}& \multirow{2}{*}{Model} &  \multirow{2}{*}{EMR}  & \multicolumn{8}{c}{Recall by class}                          \\
    \cmidrule{5-12}
                             &                           &          & & NoDam. & Crack  & Spall. & Effl. & Expos. & Rust & Scal. & Other  \\ 
    \midrule
    \parbox[t]{2mm}{\multirow{9}{*}{\rotatebox[origin=c]{90}{CODEBRIM}}}  & HO       & RN   & 63.77 & 89.33 & 77.33 & 76.67 & 62.42 & 78.67 & 80.00 & -       & -       \\
    & HO       & EN   & 59.97 & 84.00 & 72.67 & 85.33 & 53.69 & 92.00 & 80.67 & -       & -       \\
    & HO       & MN   & 63.92 & 88.00 & 74.00 & 80.67 & 68.46 & 82.00 & 70.67 & -       & -       \\
    & \textbf{HTA}      & \textbf{RN}   & \textbf{73.73} & 94.67 & \textbf{88.00} & 84.00 & \textbf{75.84} & 88.67 & 79.33 & -       & -       \\
    & HTA      & EN   & 65.35 & 90.67 & 76.67 & 87.33 & 65.10 & \textbf{92.67} & 81.33 & -       & -       \\
    & HTA      & MN   & 69.46 & 94.00 & 78.67 & 83.33 & 69.80 & 88.67 & 76.67 & -       & -       \\
    & DHA      & RN   & 70.57 & \textbf{95.33}&  87.33 & 86.67 & 75.17 & 90.00 & 82.00 & -       & -       \\
    & DHA      & EN   & 68.67 & 92.00 & 83.33 & \textbf{88.67} & 74.50 & \textbf{92.67} & \textbf{85.33} & -       & -       \\
    & DHA      & MN   & 69.15 & 90.00 & 79.33 & 81.33 & 69.13 & 86.00 & 78.67 & -       & -       \\
    \midrule
    \parbox[t]{2mm}{\multirow{9}{*}{\rotatebox[origin=c]{90}{MCDS}}}     & HO       & RN   & 44.44 & 66.67 & 63.33 & 73.33 & 63.33 & 36.67 & 75.00 & 23.33 & 23.33 \\
    & HO       & EN   & 30.37 & 33.33 & 53.33 & 43.33 & 56.67 & 10.00 & 56.67 & 0.00  & 3.33  \\
    & HO       & MN   & 47.04 & 56.67 & 70.00 & 56.67 & 70.00 & 25.00 & \textbf{76.67} & 50.00 & 36.67 \\
    & HTA      & RN   & 43.70 & 73.33 & 53.33 & 63.33 & 60.00 & 25.00 & 65.00 & \textbf{76.67} & 10.00 \\
    & HTA      & EN   & 46.30 & 46.67 & 76.67 & 55.56 & 80.00 & 26.67 & 68.33 & 30.00 & 26.67 \\
    & \textbf{HTA}      & \textbf{MN}   & \textbf{54.44} & 70.00 & 76.67 & 58.89 & \textbf{90.00} & 21.67 & 68.33 & 43.33 & \textbf{46.67} \\
    & DHA      & RN   & 48.15 & 66.67 & 73.33 & 44.44 & 86.67 & 23.33 & 65.00 & 36.67 & 43.33 \\
    & DHA      & EN   & 51.85 & 46.67 & 73.33 & \textbf{61.11} & 80.00 & \textbf{38.33} & 75.00 & 43.33 & \textbf{46.67} \\
    & DHA      & MN   & 51.85 & \textbf{80.00} & \textbf{83.33} & 57.78 & 73.33 & 30.00 & 66.67 & 33.33 & 36.67 \\                         
    \bottomrule
    \end{tabular}
    \caption{{EMR} and {recall by class} for all models and transfer learning approcheas. Only the best model from 5 runs is reported.}
    \label{tab:BestModelMetrics}
\end{table*}

\begin{figure}[h]
    \centering
    \includegraphics[width=1\linewidth]{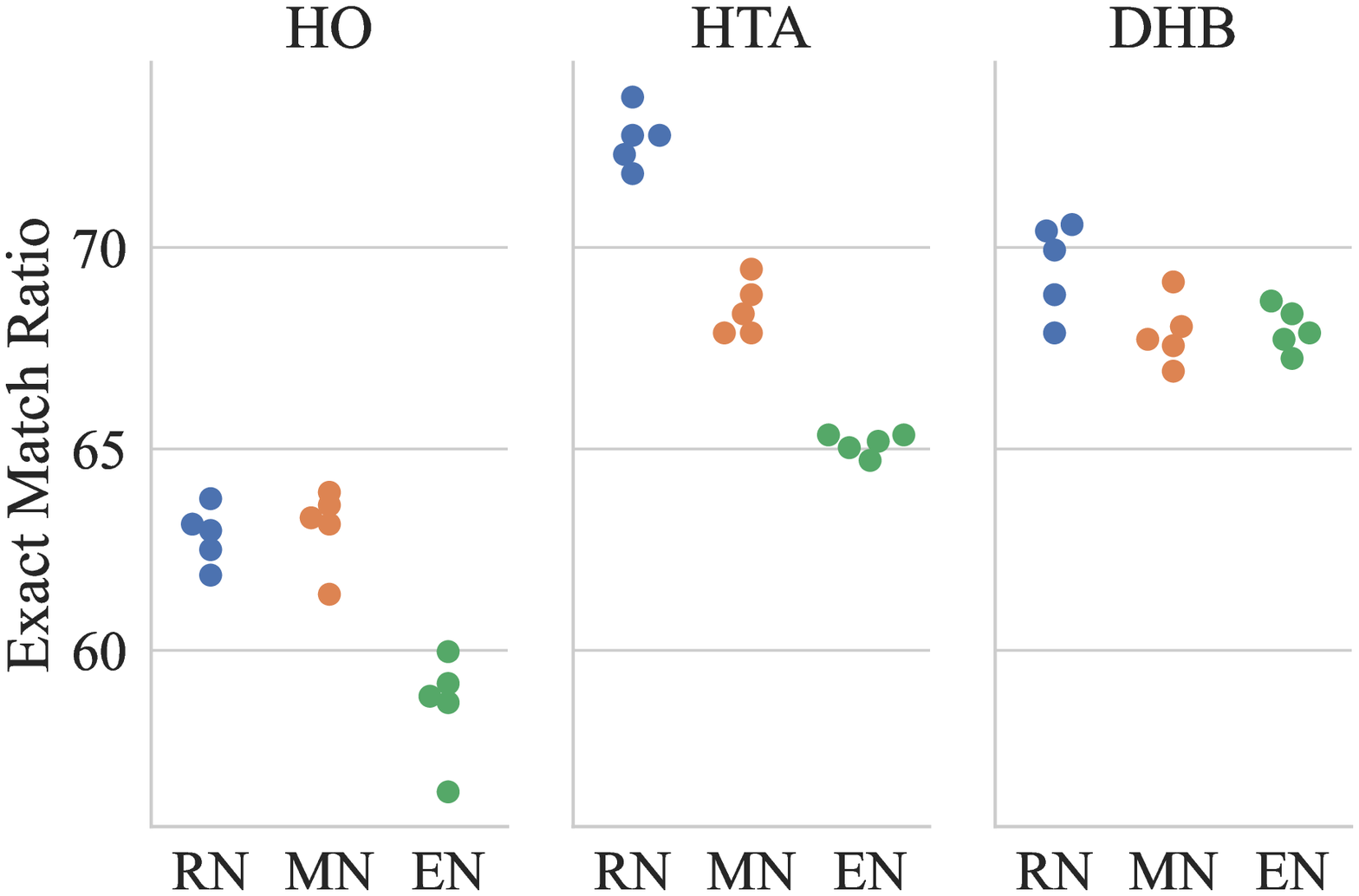}
    \caption{EMR of five runs with different seeds for CODEBRIM dataset. The used model setups are described in \cref{tab:BestHOSetting}  for \textit{HO} and in  \cref{tab:HTA_DHB} \textit{HTA} and \textit{DHB}.}
    \label{fig:ExactMatchRatioCODE}
\end{figure}

\begin{figure}[h]
    \centering
    \includegraphics[width=1\linewidth]{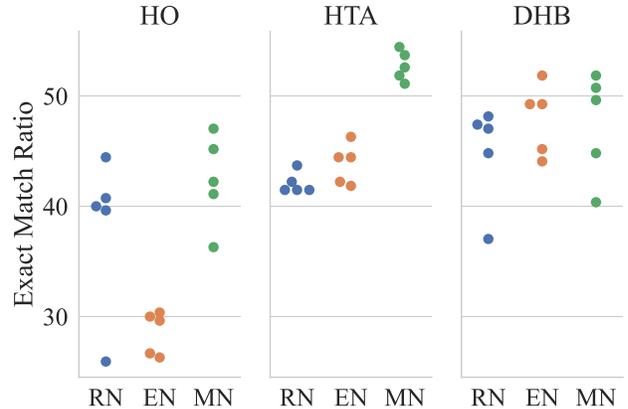}
    \caption{EMR of five runs with different seeds for MCDS dataset. The used model setups are described in \cref{tab:BestHOSetting}  for \textit{HO} and in  \cref{tab:HTA_DHB} \textit{HTA} and \textit{DHB}.}
    \label{fig:ExactMatchRatioMCDS}
\end{figure}

\color{blue} \normalsize
\color{black} 

We want to share strong baseline models for CODEBRIM as well as MCDS and therefore, went through an expensive evaluation process.
We run a hyperparameter search for three state-of-the-art architectures and investigate three different transfer learning strategies. We use \textit{EMR} as our main metric, but also provide \textit{recall per class} for a more detailed explanation for applicants. 
To ensure that our results are reliable, we repeat each final hyperparameter setting five times.  
This allows us to understand better how trustworthy the outcomes are. 
Our results are reported in  \cref{fig:ExactMatchRatioCODE}, \cref{fig:ExactMatchRatioMCDS}, and \cref{tab:BestModelMetrics}. The best results of the five runs are made available in \cref{tab:BestModelMetrics}.

\textbf{CODEBRIM.}
We observe that CODEBRIM results are stable. For the more flexible \textit{HTA} and \textit{DHB} approaches, ResNet is the best model with approximately 4 and 1.4 percent point margin to the second best model in the respective transfer learning group. Considering \textit{HO}, ResNet is on a par with MobileNet. The overall best result shows ResNet with \textit{HTA}, which scores 73.73 percent on EMR.
This model outperformed the initially published model slightly by 1.5 percent points \cite{Mundt2019}. 
Bukhsh \textit{et al.} does not provide EMR but AUROC. They score 90 percent\footnote{They use 10 classes and splits are not published.}, whereas we score 97 percent at AUROC. 
The recall is the lowest for efflorescence (75.84 percent) and the highest for recognizing exposed bars (88.67 percent). 
The current best CODEBRIM model still missed 1/4 of efflorescence damages, which is probably related to the small number of images.
On the other hand, results for exposed bars are already in a good range.

\textbf{MCDS.}
MCDS results are less distinct, as the dispersion for some settings in \cref{fig:ExactMatchRatioMCDS} indicates. 
This may be primarily due to the small dataset size. 
As for CODEBRIM, the more flexible transfer learning approaches perform better. 
The best model is MobileNet with \textit{HTA} transfer learning and it has a 3.5 percentage point lead over the second best model.
With a top EMR of 54.44 percent, performance on this dataset is substantially worse than for CODEBRIM, which is again due to the dataset size.
Recall is best for efflorescence, where 9 out of 10 damages are detected. The lowest recall is obtained for exposed bars, which have the second smallest number of images in MCDS. 
Since the original \cite{Huthwohl2019} and modified setting \cite{Bukhsh2021} are structured differently and no fix splits are provided, we cannot compare results to former research.

\textbf{Transfer learning.}
Our results show that not only training of the head but also the base part of the architecture clearly improves transfer learning results. We have included these strategies because RCD images differ a lot from ImageNet objects which were used for the pre-training of our models. There can be no conclusive assessment of which strategy performs better since we only evaluated this on two very specific datasets. 
This leads to the conclusion that, after the classical \textit{HO} approach has been conducted, it is reasonable to train all weights. \textit{DHB} has the advantage of requiring only one pass.

%-------------------------------------------------------------------------
\section{Conclusions}
\label{sec:Conclusion}

The building inspection toolkit simplifies access to datasets and introduces splits in case they are missing. 
It builds the base for a unified evaluation process by providing metrics from the machine
learning and the application perspective. Currently, EMR and recall by class are used. 
We also provide baseline models trained with three different transfer learning strategies. 
An online leaderboard finally facilitates comparability between models and accelerates the progress within this application oriented research area.
Experiments on the CODEBRIM dataset showed that ResNet works best, whereas MobileNet with \textit{HTA} is the best model for MCDS. 
During experiments we found that it is of great importance not only to train parameters in the model’s head but also to update the filters in the base model. 
This is relevant because the pre-trained models were trained with images from a different domain and, therefore, it is beneficial if filters can be adjusted as well.

All parts of the building inspection toolkit will help to speed up research in RCD damage recognition and give it a higher priority and greater visibility.
In future work, focus will be on combined occurrences of several damages, like spalling with exposed reinforcement and rust staining.
It is planned to (\textit{i}) extend data augmentation methods, like manipulating color space, injecting noise or mixing images \cite{Shorten2019}, (\textit{ii}) considering models that can attend to different damages and (\textit{iii}) label correlations through a graph superimposing framework  \cite{Wang2020}.

%%%%%%%%% REFERENCES
\newpage

{\small
\bibliographystyle{ieee_fullname}
\bibliography{egbib, library}
}

%%%%%%%%% Appendix
\appendix
\noindent

%-------------------------------------------------------------------------
\section{Data Augmentation}
\label{app:DataAugmentation}

The data augmentation process of all training sets follow the subsequent steps:
    \begin{enumerate}[noitemsep]
        \item Resizing to 1.1 ($\approx1/0.875$) times the input resolution 224 %\cite{pytorch-tut2021}
        (cropping the central 87.5$\%$ is common for testing on ImageNet \cite{Kornblith2019})
        \item Center-cropping to a $224\times224$ image
        \item Randomly rotating up to 30 degree
        \item Flipping horizontally and vertically with 50\% probability
        \item Normalizing with the mean and standard deviation from ImageNet 
    \end{enumerate}

The preprocessing of the validation and testing data involved resizing, center-cropping and normalization exclusively.

\end{document}